# Why & When Deep Learning Works: Looking Inside Deep Learning

**Ronny Ronen**  ronny.ronen@intel.com
*The Intel Collaborative Research Institute for Computational Intelligence (ICRI-CI)[1]*

In recent years, Deep Learning has emerged as the leading technology for accomplishing broad range of artificial intelligence tasks (LeCun et al. (2015); Goodfellow et al. (2016)). Deep learning is the state-of-the-art approach across many domains, including object recognition and identification, text understating and translation, question answering, and more. In addition, it is expected to play a key role in many new usages deemed almost impossible before, such as fully autonomous driving.

While the ability of Deep Learning to solve complex problems has been demonstrated again and again, there is still a lot of mystery as to why it works, what is it really capable of accomplishing, and when it works (and when it does not). Such an understanding is important for both theoreticians and practitioners, in order to know how such methods can be utilized safely and in the best possible manner. An emerging body of work has sought to develop some insights in this direction, but much remains unknown. The general feeling is that Deep learning is still by and large "black magic" we know it works, but we do not truly understand why. This lack of knowledge disturbs the scientists and are a cause for concern for developers would you let an autonomous car be driven by a system whose mechanisms and weak spots are not fully understood?

The Intel Collaborative Research Institute for Computational Intelligence (ICRI-CI) has been heavily supporting Machine Learning and Deep Learning research from its foundation in 2012. We have asked six leading ICRI-CI Deep Learning researchers to address the challenge of "Why & When Deep Learning works", with the goal of looking inside Deep Learning, providing insights on how deep networks function, and uncovering key observations on their expressiveness, limitations, and potential.

The output of this challenge call was quite impressive, resulting in five papers that address different facets of deep learning. These papers summarize the researchers' ongoing recent work published in leading conferences and journals as well as new research results made especially for this compilations. These different facets include a high-level understating of why and when deep networks work (and do not work), the impact of geometry on the expressiveness of deep networks, and making deep networks interpretable.

**Understating of why and when deep networks work (and do not work)**

1. **Naftali Tishby and Ravid Schwartz-Ziv** in **Opening the Black Box of Deep Neural Networks via Information** study Deep Networks by analyzing their information-theoretic properties, looking at what information on the input and output each layer preserves, and suggests that the network implicitly attempts to optimize the Information-Bottleneck (IB) tradeoff between compression and prediction, successively, for each layer. Moreover, they show that the stochastic gradient descent (SGD) epochs used to train such networks have two

---

[1] This work was done with the support of the Intel Collaborative Research institute for Computational Intelligence (ICRI-CI). This paper is the preface part of the 'Why & When Deep Learning works looking inside Deep Learning' ICRI-CI paper bundle.

distinct phases for each layer: fast empirical error minimization, followed by slow representation compression. They then present a new theoretical argument for the computational benefit of the hidden layers

2. **Shai Shalev-Shwartz, Ohad Shamir and Shaked Shamma** in [Failures of Gradient-Based Deep Learning](#) attempt to gain a deeper understanding of the difficulties and limitations associated with common approaches and algorithms. They describe four families of problems for which some of the commonly used existing algorithms fail or suffer significant difficulty, illustrate the failures through practical experiments, and provide theoretical insights explaining their source and suggest remedies to overcome the failures that lead to performance improvements.

**The impact of geometry on the expressiveness of deep networks**

3. **Amnon Shashua, Nadav Cohen, Or Sharir, Ronen Tamari, David Yakira and Yoav Levine** in [Analysis and Design of Convolutional Networks via Hierarchical Tensor Decompositions](#) analyze the expressive properties of deep convolutional networks. Through an equivalence to hierarchical tensor decompositions, they study the expressive efficiency and inductive bias of various architectural features in convolutional networks (depth, width, pooling geometry, inter-connectivity, overlapping operations etc.). Their results shed light on the demonstrated effectiveness of convolutional networks, and in addition, provide new tools for network design.

4. **Nathan Srebro, Behnam Neyshabur, Ryota Tomioka and Ruslan Salakhutdinov** in [Geometry of Optimization and Implicit Regularization in Deep Learning](#) argue that the optimization methods used for training neural networks play a crucial role in generalization ability of deep learning models, through implicit regularization. They demonstrate that generalization ability is not controlled simply by network size, but rather by some other implicit control. Then, by studying the geometry of the parameter space of deep networks and devising an optimization algorithm attuned to this geometry, they demonstrate how changing the empirical optimization procedure can improve generalization performance.

**Interpretability of deep networks**

5. **Shie Mannor, Tom Zahavy and Nir Baram** in [Graying the black box: Understanding DQNs](#) present a methodology and tools to analyze Deep Q-networks (DQNs) in a non-blind matter. They propose a new model, the Semi Aggregated Markov Decision Process (SAMDP), and an algorithm that learns it automatically. Using these tools they reveal that the features learned by DQNs aggregate the state space in a hierarchical fashion, explaining its success. Moreover, they are able to look into the network to understand and describe the policies learned by DQNs for three different Atari2600 games and suggest ways to interpret, debug and optimize deep neural networks in reinforcement learning.